\documentclass[conference]{IEEEtran}
\IEEEoverridecommandlockouts
\usepackage{cite}
\usepackage{amsmath,amssymb,amsfonts}
\usepackage{algorithmic}
\usepackage{graphicx}
\usepackage{textcomp}
\usepackage{xcolor}
\usepackage{algorithm}
\usepackage{algorithmic}
\usepackage{comment}
\usepackage[hyphens]{url}        
\usepackage[breaklinks]{hyperref} 
\usepackage{subfigure}

\def\BibTeX{{\rm B\kern-.05em{\sc i\kern-.025em b}\kern-.08em
    T\kern-.1667em\lower.7ex\hbox{E}\kern-.125emX}}
\begin{document}

\title{DYNAMAX: Dynamic computing for Transformers and Mamba based architectures\\
}


\author{
\IEEEauthorblockN{Miguel Nogales$^{1}$, Matteo Gambella$^{2}$, and Manuel Roveri$^{2}$}
\IEEEauthorblockA{$^{1}$ Università della Svizzera Italiana, Lugano, Switzerland \\
$^{2}$ Politecnico di Milano, Milano, Italy\\
Email: miguel.nogales@usi.ch, \{matteo.gambella, manuel.roveri\}@polimi.it}
}


\maketitle
\begin{abstract}
Early exits (EEs) offer a promising approach to reducing computational costs and latency by dynamically terminating inference once a satisfactory prediction confidence on a data sample is achieved. Although many works integrate EEs into encoder-only Transformers, their application to decoder-only architectures and, more importantly, Mamba models, a novel family of state-space architectures in the LLM realm, remains insufficiently explored. This work introduces DYNAMAX, the first framework to exploit the unique properties of Mamba architectures for early exit mechanisms. We not only integrate EEs into Mamba but also repurpose Mamba as an efficient EE classifier for both Mamba-based and transformer-based LLMs, showcasing its versatility. Our experiments employ the Mistral 7B transformer compared to the Codestral 7B Mamba model, using data sets such as TruthfulQA, CoQA, and TriviaQA to evaluate computational savings, accuracy, and consistency. The results highlight the adaptability of Mamba as a powerful EE classifier and its efficiency in balancing computational cost and performance quality across NLP tasks. By leveraging Mamba's inherent design for dynamic processing, we open pathways for scalable and efficient inference in embedded applications and resource-constrained environments. This study underscores the transformative potential of Mamba in redefining dynamic computing paradigms for LLMs.
\end{abstract}

\begin{IEEEkeywords}
Large Language Models (LLMs), Dynamic Computing, Early Exit Neural Networks, Transformers, State-Space models
\end{IEEEkeywords}

\section{Introduction}
\label{sec:introduction}

Natural Language Processing (NLP) has experienced transformative growth due to advancements in neural network architectures, particularly decoder-only models that leverage autoregressive processes for text generation \cite{success_NLP_1,success_NLP_2_fix}. At the core of these innovations lie Transformer-based architectures, which utilize self-attention mechanisms to enable large-scale parallel processing \cite{transformer_fix}. This paradigm shift has substantially enhanced capabilities in language understanding and generation, setting new benchmarks for generalization and performance in NLP tasks (and many others). 

Decoder-only models have emerged as key contributors to this progress. Open-source frameworks like Llama \cite{llama} and Mistral~\cite{mistral7b}, alongside proprietary systems such as GPT \cite{gpt2} and Claude \cite{claude3}, demonstrate the effectiveness of pre-training on large-scale corpora followed by task-specific fine-tuning. This transfer learning approach enables these models to generalize across domains, delivering state-of-the-art results in text generation, summarization, and other language-related applications.
Despite these advancements, the computational and energy demands of Transformers remain a significant challenge, especially since their model size is not only steadily growing, but also the inference-time computation needs have increased due to the use of Test Time Compute (TTC)\cite{ttc}, where models such as closed OpenAI's o1 \cite{o1} or open source DeepSeek's R1 \cite{R1} make use of it to enhance performance. To address this issue, this work explores technical innovations aimed at improving efficiency while maintaining high performance. Architectures such as Mamba \cite{mamba}, which integrate state-space models, offer a novel solution to mitigate the scalability limitations inherent to standard Transformers. Furthermore, emerging techniques such as early exit (EE) mechanisms \cite{eeoriginal_fix}, widely used in computer vision \cite{branchynet_fix, skipnet}, are being adapted to NLP \cite{dynabert_fix}. By embedding auxiliary classifiers within intermediate layers, these mechanisms allow models to terminate processing early when confidence thresholds are met, reducing computational overhead and energy consumption.


To the best of our knowledge,
there is no prior study on the performance of Mamba with EEs, nor EEs classifiers based on Mamba architectures.

In this paper, we present DYNAMAX, a framework of dynamic computing for decoder-only Transformers and Mamba-based architectures with EEs aiming at improving the trade-off between the performance and computational cost of these LLMs. Early exit classifiers are added to the latter half of the LLM and trained by knowledge distillation from the full model and applying a relaxation to avoid training instability. The inference follows a general token forwarding scheme with EEs where it is possible to stop computation when enough confidence is achieved. In Transformers, a simple tweak in how missing states are handled permits more enhancement of the computational efficiency. Three evaluation datasets have been selected to assess the effectiveness of the framework on diverse linguistic tasks. In particular, we compared the efficacy in increasing computational efficiency of EE techniques with respect to another method named layer pruning, which showed to be promising for LLMs \cite{ineffect}.

The innovations presented in this work are the following:

\begin{itemize}
    \item The addition of EEs to Mamba architectures.
    \item The use of Mamba as an EE classifier exploiting its unique properties.
    \item An alternative way of training the EE classifiers.
    \item A different, more efficient, way to deal with missing states in Decoder-only Transformers.
\end{itemize}

The work is structured as follows: following the Introduction, section \ref{sec:background} will cover technicalities about the models' architectures, early exits, and current implementations. Section \ref{sec:solution} will cover the implementations of the innovations presented in the work. Lastly, the results of the conducted experimental campaign will be presented, showing how computation savings affect performance. To facilitate comparisons and reproducibility, the source code of DYNAMAX is released to the scientific community as a public repository. \footnote{\url{https://github.com/Xigm/DYNAMAX}}


\section{Related literature}

The usage of EEs in Transformers has been widely addressed with encoder-only models (such as BERT \cite{bert}), where they have found great success. Some examples of this are \cite{adavit} or \cite{lgvit_fix}, where EEs accelerate the costly inference of those models. Additionally, the work \cite{nodeformer_fix} satisfactorily applies EEs to Transformers in graph neural networks for additional speed-up gains. There are also examples of Transformers with EEs in vision applications such as \cite{cascadebert} or \cite{fastbert}. In the field of Transformers which include decoders with EEs attached, the works \cite{copyingstate} and \cite{calm_fix} are found, which set the basis of EE-accelerated Transformer-based models. Their performance, while modest, opens up a new framework for the inference of LLMs. Lastly, focusing on decoder-only models, the works \cite{robustee} and \cite{eellm} found great speed-ups by using draft models to generate the text. These approaches focus on the speed-up aspect of EEs, but do not tackle the computational complexity or energy spent.


Key works on decoder models with EEs include \cite{calm_fix} and \cite{copyingstate}, where the latter introduced the concept and CALM refined its framework. EEs are applied across all model blocks, making token-wise exit decisions based on entropy, state saturation, and neural network-based predictions. Entropy-based exits offer high confidence but incur high computational costs due to Softmax operations. State saturation, which measures hidden state differences, is simpler but less effective. Neural network predictors provide a balanced trade-off between efficiency and accuracy. Unlike other EE methods, these frameworks share a common output classifier across all exits.



There are latter works such as \cite{robustee}, which introduce the FREE framework, demonstrating competitive results against CALM. Like prior research, they utilized T5 family models \cite{t5_fix}, employing a dual-model approach where a shallow model partially computes tokens via early exits, while a deep model processes them fully. Their key contribution is an innovative decoding strategy: a new cache mechanism retains early-exited tokens, enabling parallel computation of their key values when full processing is required.

Similarly, the work by Chen et al. \cite{eellm} explores a comprehensive framework for training and inference of LLMs with EE. Achieving speed-ups of up to $\times3$, their approach leverages extensive parallelism, including pipeline parallelism during decoding. By fully utilizing GPU resources, they enable the use of entire encoder transformer models as early exit classifiers, significantly enhancing efficiency.

In this work, from the three main methods to assign a confidence measure to exit in decoders with EEs, the neural network-based one criterion is selected. This method is more suited to fulfill the requirements imposed for constrained environments, while maintaining performance.

\section{Technical background}
\label{sec:background}

\subsection{Transformer and Mamba Architectures}

The Transformer \cite{transformer_fix} and Mamba \cite{mamba} model architectures represent two distinct approaches to handling long-range dependencies in sequential data. Transformers leverage self-attention to capture global context within sequences by computing pairwise attention across all tokens. Self-attention computes a weighted sum of values ($V$), where the weights are determined by the compatibility between the query ($Q$) and key ($K$) vectors. Mathematically:
\[
\text{Attention}(Q, K, V) = \text{softmax}\left(\frac{QK^\top}{\sqrt{d_k}}\right)V,
\]
where $d_k$ is the dimension of the key vectors.
The multi-head attention mechanism extends this by computing multiple attention heads in parallel:
\[
\text{MultiHead}(Q, K, V) = \text{Concat}(\text{head}_1, \ldots, \text{head}_h)W^O,
\]
where each attention head is computed as:
\[
\text{head}_i = \text{Attention}(QW_i^Q, KW_i^K, VW_i^V).
\]

Transformer models offer powerful parallelization and expressive capabilities but are computationally and memory-intensive, with training costs scaling quadratically with sequence length. To enhance efficiency, transfer learning has become a key strategy in NLP, enabling models to be pre-trained on large datasets and fine-tuned for specific tasks. While pretraining is resource-intensive, parameter-efficient fine-tuning techniques (PEFT) \cite{peft}, such as LoRA (low-rank adaptation) \cite{lora} or adapters, reduce computational demands, allowing models to leverage open-source pre-trained weights for adaptation across domains.

The main focus of this work is the reduction of computational complexity; the number of operations of the Transformer block is re-examined. The main source of computational cost of the block is the self-attention mechanism and the feedforward network. Note that we will not take into account the layer norm or the residual connections' compute. 

\begin{equation}
\begin{aligned}
\text{Total Operations} &= \text{Total}_{\text{MHSA}} + \text{Total}_{\text{FFN}} \\
&= \left[ 8 \times T \times d_{\text{model}}^2 + 4 \times T^2 \times d_{\text{model}}\right] \\
&\quad \quad + \left[ 16 \times T \times d_{\text{model}}^2\right]  \\
&= \left[ 24 \times T \times d_{\text{model}}^2 + 4 \times T^2 \times d_{\text{model}} \right]
\end{aligned}
\label{eq:cost_block}
\end{equation}

$T$ represents the sequence length and $d_{model}$ represents the internal dimension of the Transformer. We can see how attention contributes with quadratic dependence on the length. The complexity of inference implementing one token forwarding with KV caching would be very similar, but dependence on the sequence length is decreased in all terms by one order of magnitude. 

However, Mamba is an efficient architecture that utilizes state-space models (SSMs) for sequence processing. Unlike Transformers, which rely on self-attention, Mamba processes sequences using linear operations in combination with learned dynamics.
SSMs describe the evolution of a hidden state $x_t$ over time based on input $u_t$:
\[
x_{t+1} = Ax_t + Bu_t, \quad y_t = Cx_t + Du_t,
\]
where $A$, $B$, $C$, and $D$ are learnable parameters, $x_t$ is the state vector, $u_t$ is the input, and $y_t$ is the output.

The convolutional form of SSMs allows for efficient sequence processing:
\[
y = \text{SSM}(u) = C \ast (K \ast u) + Du,
\]
where $K$ is a kernel derived from the state-space dynamics. 

Mamba exploits SSMs to capture input-dependent information, trading some parallelization for scalability in long-sequence tasks. To address this, a hardware-aware algorithm ensures efficient training. Mamba 2 \cite{mamba2_fix} introduces architectural optimizations, enhancing training flexibility and connecting SSMs with linear attention techniques. Unlike self-attention, which explicitly computes global dependencies, Mamba captures them implicitly through learned dynamics with a constant-sized state.

In a similar way as the Transformer, the computational complexity can be accurately approximated by the cost of its projection matrices (the input and output projections of the Mamba block) because their other components, such as the one-dimensional convolution and the SSM block, are at least an order of magnitude smaller.

\begin{equation}
\begin{aligned}
\text{Total Operations} &= \text{Total}_{\text{input}} + \text{Total}_{\text{output}} \\
&= \left[ 2 \times d_{\text{model}}^2 + 2 \times d_{\text{model}}^2 \right] \\
&\quad \quad + \left[ 2 \times n_{\text{groups}}  \times d_{\text{state}} \times d_{\text{model}} \vphantom{d_{\text{model}}^2} \right] \\
&\quad \quad + \left[ 2 \times d_{\text{model}}^2 \right]  \\
&= \left[ 6 \times d_{\text{model}}^2 + 2 \times n_{\text{groups}} \times d_{\text{state}} \times d_{\text{model}} \right]
\end{aligned}
\label{eq:cost_block_mamba}
\end{equation}

The first term of Equation \ref{eq:cost_block_mamba} is the projection of input data, precisely the computation of matrices, the projection of the input and computation of matrices B and C. The output term is just the output projection of the computed state.

\subsection{Early Exit Mechanisms}

Early exit mechanisms (EE) \cite{eeoriginal_fix} enhance efficiency by halting computations early when confidence thresholds are met, reducing resource usage and processing time. This is particularly beneficial for large models like Transformers, where increasing sequence length amplifies computational demands, enabling faster predictions without compromising performance. In real-time applications, EEs dynamically allocate resources based on data complexity, which is advantageous for models like SSMs that benefit from linear complexity.

EE mechanisms function via checkpoints at various model stages, assessing confidence in intermediate predictions and terminating computation when thresholds are met. Typically, checkpoints are positioned after each block in architectures like Transformers or SSM-based models (e.g., Mamba), allowing precise resource management. Training methods such as joint training \cite{edanas}, layer-wise training \cite{greedylayerwise}, and knowledge distillation \cite{destill} enhance intermediate layer performance, enabling confident early predictions.

Another important implication of EEs is that they are orthogonal to any other efficiency-related modification of NNs. The most important techniques in this category are pruning\cite{surveypruning_fix}, which removes redundant weights or neurons to improve efficiency with minimal performance loss, quantization\cite{1bitllm}, which reduces the precision of the weights (e.g. from Float16 to Float8) and efficient attention techniques, such as works \cite{linformer, longformer}, where they modify the attention mechanism to reduce its complexity, whether it's by linearizing the operation (removing the softmax operation) or by using windowed attention. The only one which is not that suitable is batching, the processing of different data samples in parallel, which even though the computation reduction is achieved, does not receive the benefits of reduced delay. In the context of this
work, layer pruning \cite{ineffect} is particulary relevant, as it allows for direct comparison with EEs. Both techniques make the model use a smaller number of layers than the original one, the main difference being that EE performs the allocation of computation on a per-token basis. Instead, layer pruning assigns a static amount of block for all tokens. This allows reducing memory footprint, unlike EEs. 

\begin{algorithm}[t]
\caption{Token Forwarding with Early Exits during Inference in Large Language Models}
\label{alg:token_forwarding}
\begin{algorithmic}[1]
    \STATE \textbf{Input:} Token $T$, Early Exit Threshold $\theta$
    \STATE \textbf{Output:} Final output $output$
    
    \STATE $x \gets \text{tokenize}(T)$
    \STATE $pe \gets \text{positional\_encoding}(x)$
    \STATE $z \gets x + pe$
    
    \FOR{each transformer block $b$ in the model backbone}
        \STATE $z \gets b(z)$
        \IF{$b$ contains an early exit classifier}
            \STATE $exit\_output \gets \text{early\_exit}(z)$
            \IF{$exit\_output \geq \theta$} 
                \FOR{each subsequent layer $b'$ in the remaining model backbone}
                    \STATE $states \gets \text{partial\_forward}(z)$
                \ENDFOR
                \STATE \textbf{break} 
            \ENDIF
        \ENDIF
    \ENDFOR
    
    \STATE $z \gets \text{last\_block}(z)$
    \STATE $z \gets \text{norm}(z)$
    \STATE $output \gets \text{head}(z)$
    \STATE \RETURN $output$
\end{algorithmic}
\end{algorithm}

\section{Proposed solution and implementation}\label{sec:solution}

In this section, we present the technical decisions and approaches used in the implementation of early exit mechanisms in LLMs. This section details the selection of models and datasets, as well as the training and inference strategies for implementing early exits in Transformer and Mamba architectures.

\subsection{Models}

Pre-trained weights for Mistral 7B v0.3 \cite{mistral7b} and Codestral Mamba 7B \cite{codestral}, similar in model size, available in the Hugging Face repository \cite{hf}, were used for this study. Mistral 7B is a Transformer, set with pre-trained weights, which integrates its architecture with sliding window and grouped query attention to optimize computational cost, while Codestral Mamba 7B, specifically optimized for code tasks, employs Mamba 2 blocks to improve efficiency in complex tasks. These models serve as a baseline to analyze the impact of early exits on computational efficiency and overall performance.

\subsection{Early Exits for Computational Efficiency in Autoregressive Generation}

Early exits have been implemented to reduce the computational cost of autoregressive generation by enabling intermediate exits based on model confidence. To minimize overhead, neural network classifiers are integrated at specific layers in the latter portion of the models, as early exits at these positions are more likely to yield accurate results without disrupting token dependencies in the generation process. These EE classifiers are designed in three distinct manners:

\begin{itemize}
    \item \textbf{CALM Style}: A simple one-layer feed-forward network, mimicking the ones used with encoder-decoder Transformers in \cite{calm_fix}.
    \item \textbf{Transformer Block-Based Network (FFN)}: An feed-forward network similar to those in transformer blocks, taking inputs sized to the backbone's internal dimension and expanding to four times the size at the second layer, and then reducing the size to two, to flag the exit.
    \item \textbf{Mamba Block}: In a similar way as the prior, the Mamba block is used, modifying the output projection layer to output two values instead of the whole internal-dimension-sized hidden state.
\end{itemize}

The rationale for employing the Mamba block is its ability to account for state dependencies, which FFNs lack without compromising computational cost, which is the downside of the Transformer block. Transformer blocks are not utilized as EE classifiers due to their complexity and the associated increase in computational cost, particularly regarding the growing size of their key-value cache during forwarding. 

Each EE classifier outputs two values, processed through a Softmax function, to decide whether to exit at the current layer or continue to the next. This decision is guided by a pre-defined confidence threshold, which determines the computational budget. Higher thresholds require greater confidence from the classifier to exit early, thereby reducing computational cost less than lower thresholds but maintaining the original performance.

The classifiers are connected to the backbone by redirecting their hidden states, after applying layer normalization, to the network. In the case of the Mamba backbone, taking into account that it has a constant size state, it was considered also to include as input its state, or a projection of it in the case of the convolutional block, but it was discarded due to growth in parameter size and not much improvement in prediction quality. When using Mamba as EE classifier, the general strategy is maintained but removing the typical skip connection of the block.

\subsection{Training Early Exit Classifiers}

The training of EE classifiers follows a knowledge distillation approach, where the early exit outputs are compared to the full model output, referred to as the \textit{oracle}. Classifier inputs consist of the hidden values of the token being forwarded. Training is supervised using the cross-entropy loss, optimizing classifier accuracy against the oracle output.

To address the challenge of unbalanced targets during training, caused by the strict requirement for EE outputs to match the oracle exactly, this work proposes a relaxation inspired by the top-$k$ sampling process. Exits are triggered if the most probable EE token is among the top-$k$ most probable tokens predicted by the oracle, being equivalent to the CALM's oracle setting with $k=1$. This approach tolerates minor variations in output probabilities and enhances training stability.

Training is parallelized across classifiers. In the case of the Mamba EE classifier, we use the parallel form of the SSM to perform this training. When jointly training them, the overall loss is adjusted using linearly decaying weights based on each classifier's position within the network.




\subsection{Inference with Early Exits}

The inference process for EEs differs between Transformer and Mamba architectures due to their structural variations. In Transformers, KV caching is used to store intermediate representations essential for autoregressive generation. When an EE is triggered, the KV cache of subsequent Transformer blocks lacks representations for exited tokens. This issue is traditionally addressed by partially forwarding the token through the remaining blocks to update their states. However, this work proposes a more computationally efficient alternative: copying the cached values directly to subsequent blocks, bypassing the need for recomputation. This way, for each Transformer block $block_n$ which is not activated for the forwarding of the current token, will have its KV cache updated with the keys and value from the 


In the Mamba architecture, the recurrence-based design allows for incremental updates to internal states. When an early exit is triggered, state updates are managed similarly to the partial forward method in Transformers. Alternatively, skipped recomputation can be employed, leaving the states unchanged, under the assumption that a posterior token forwarded through the entire model will eventually update all states with any missing information. A repetition penalty is introduced to mitigate token repetition during generation. In the case of Mamba as EE classifier, also the recurrent mode is employed to allow for constant complexity, updating in each forward the classifier state. No recomputation of states is used when using Mamba as EE classifier.

The overall inference process for early exits, generalized to both Transformers and Mamba architectures, is described in Algorithm \ref{alg:token_forwarding}. This algorithm outlines the steps for token forwarding during inference, where tokens are sequentially processed through layers. At each layer, the EE condition is evaluated. If met, the loop terminates, reducing computational cost. The shared model head subsequently generates the final output, ensuring an efficient and lightweight early exit approach.

\section{Results}

This section presents the experimental results and a discussion of the effectiveness of EEs versus layer pruning in LLMs. The first part addresses model performance across selected evaluation tasks, focusing on the Mamba and Transformer models with various EE configurations and with and without recomputation. The computational reduction factor used to evaluate the models is the relationship between the total number of operations of using the whole model versus the computation actually used by the model. In the case of layer pruning, it is directly related to the number of enabled blocks. In the case of EEs, the complexity, aside from the number of backbone blocks activated, has two additional factors. First, the evaluation of the EEs, which, depending on the architecture, may contribute to the overall computation; in the case of the CALM classifiers, it is negligible, while in the case of Mamba and the FFN, it will contribute. The other source of computational cost is the recomputation of states. This process adds to the total expenditure a fraction of the cost of the entire block (partial forward). In the case of Transformers, this process consists of the computation of the KV cache with the current token, which costs $\frac{1}{6}$ of the whole block, while in Mamba it is the update of the 1-dimensional convolutional layer and the SSM for a cost of $\frac{9}{26}$. Lastly, performance is evaluated by sweeping the values of confidence thresholds. The same value is used for each classifier in a single test, to reduce the search space, as combinations of these would have resulted in too many different evaluations. Sometimes, when the threshold is set too low, the model proceeds to output inconsistent responses that are composed of the same repeated token. If this event occurs a certain number of times, that configuration for the evaluation is not considered valid and then not shown in the graphs.

\subsection{Datasets for training and benchmarking}

We use multiple datasets for training and evaluation. For training, the FineWeb-Edu dataset \cite{finewebedu} is used, comprising over one billion curated paragraphs. Only a sample of 10 billion tokens was used, focusing on representative examples that avoid repetitive patterns and enhance model training efficiency. Three evaluation datasets have been chosen to assess different language understanding tasks: TruthfulQA \cite{truthfulqa}, CoQA \cite{coqa_fix} and TriviaQA \cite{triviqa}. A crucial aspect of these datasets is that they evaluate text generation, while also accounting for knowledge. TriviaQA is a test which leverages both generation and knowledge, but many of them consist of single-word answers. For that reason, a subset of one thousand samples was selected from the ones with the longer answers. In this way, the knowledge-based aspect of the test is achieved while improving the text generation part. The metric used to evaluate this test is exact match. CoQA emphasizes conversational question-answering, testing the ability of the model to maintain context. Its main metric is also an exact match, but F1 is also provided. Lastly, TruthfulQA for text generation is used, evaluated with BLEU and Rouge metrics. This set of datasets provides a comprehensive evaluation framework for assessing the effectiveness of early exit in various linguistic tasks. 

\subsection{Performance in the Tasks}

We used procedures available from Eleuther AI for the evaluation harness \cite{evalharness}.
In the following figures, we compare the performance of early exits versus layer pruning techniques, using task-specific metrics: exact match for TriviaQA, exact match and F1 for CoQA, and Bleu and Rouge for TruthfulQA. It is worth noting that these metrics, specially the ones for TriviaQA and CoQA, provide a strict evaluation for both text generation and knowledge performance of the models. TriviaQA is evaluated with two shots, while TruthfulQA is evaluated with one. The upper right corner of each plot represents optimal performance, achieving high accuracy with maximum computational savings. Typically, model performance decreases as computational resources are reduced, moving from the top left to the bottom right of each figure. In these experiments, a configuration of four early exits is used, placed in the second half of the model's backbone. Layer pruning is implemented by
disabling from the $Nth$ - 1 block, as many blocks as desired, following work \cite{ineffect}. No afterwards healing of the backbone is performed, considering both the inclusion of EE's and layer pruning an adhoc method. 

\subsubsection{TriviaQA}

\begin{figure}[h]
    \centering
    \begin{subfigure}
        \centering
        \includegraphics[width=0.5\textwidth, trim={0 0 0 0}, clip]{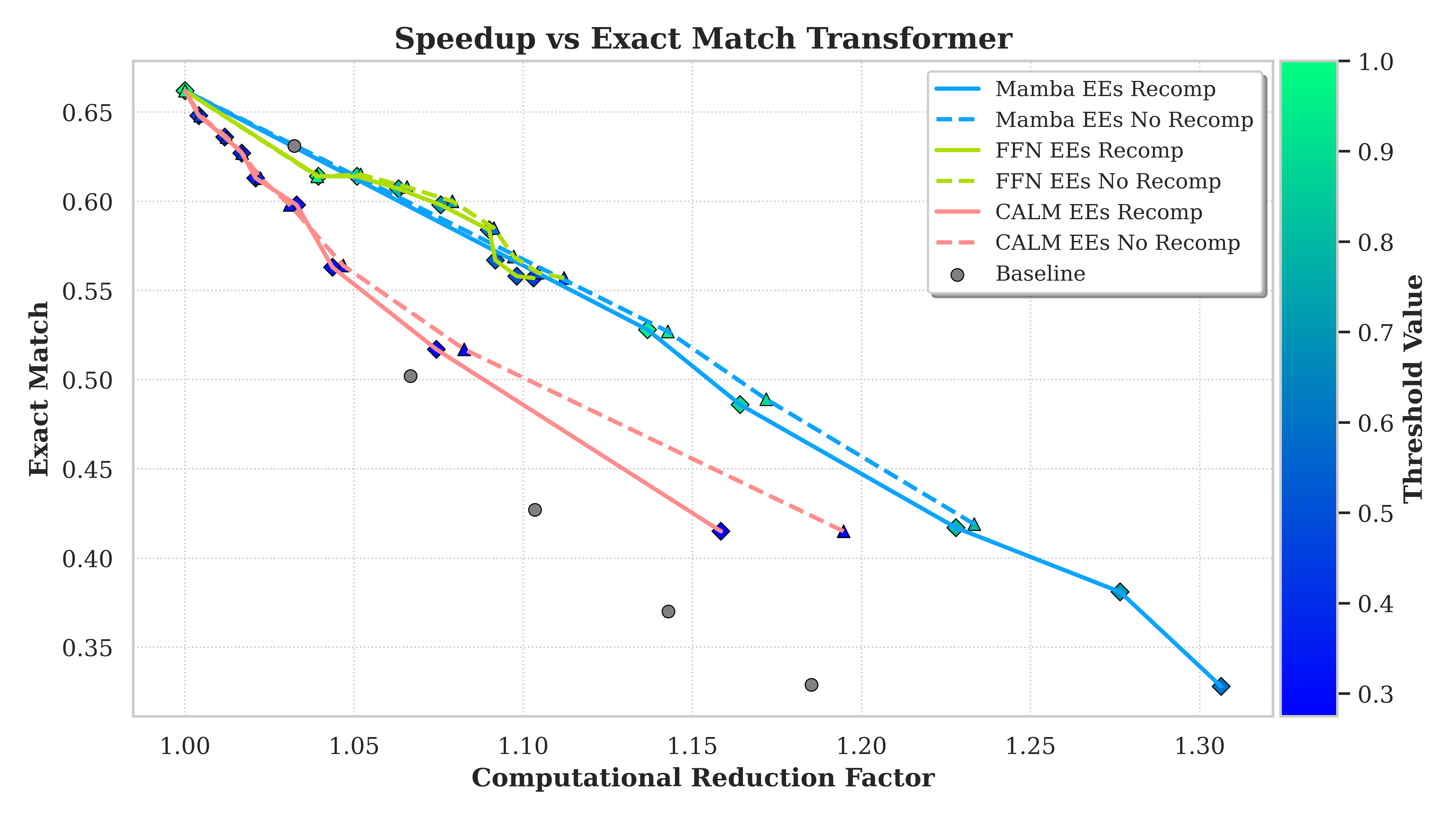}
        \label{fig:mistral_triviaqa}
    \end{subfigure}
    \hfill
    \begin{subfigure}
        \centering
        \includegraphics[width=0.5\textwidth, trim={0 0 0 0}, clip]{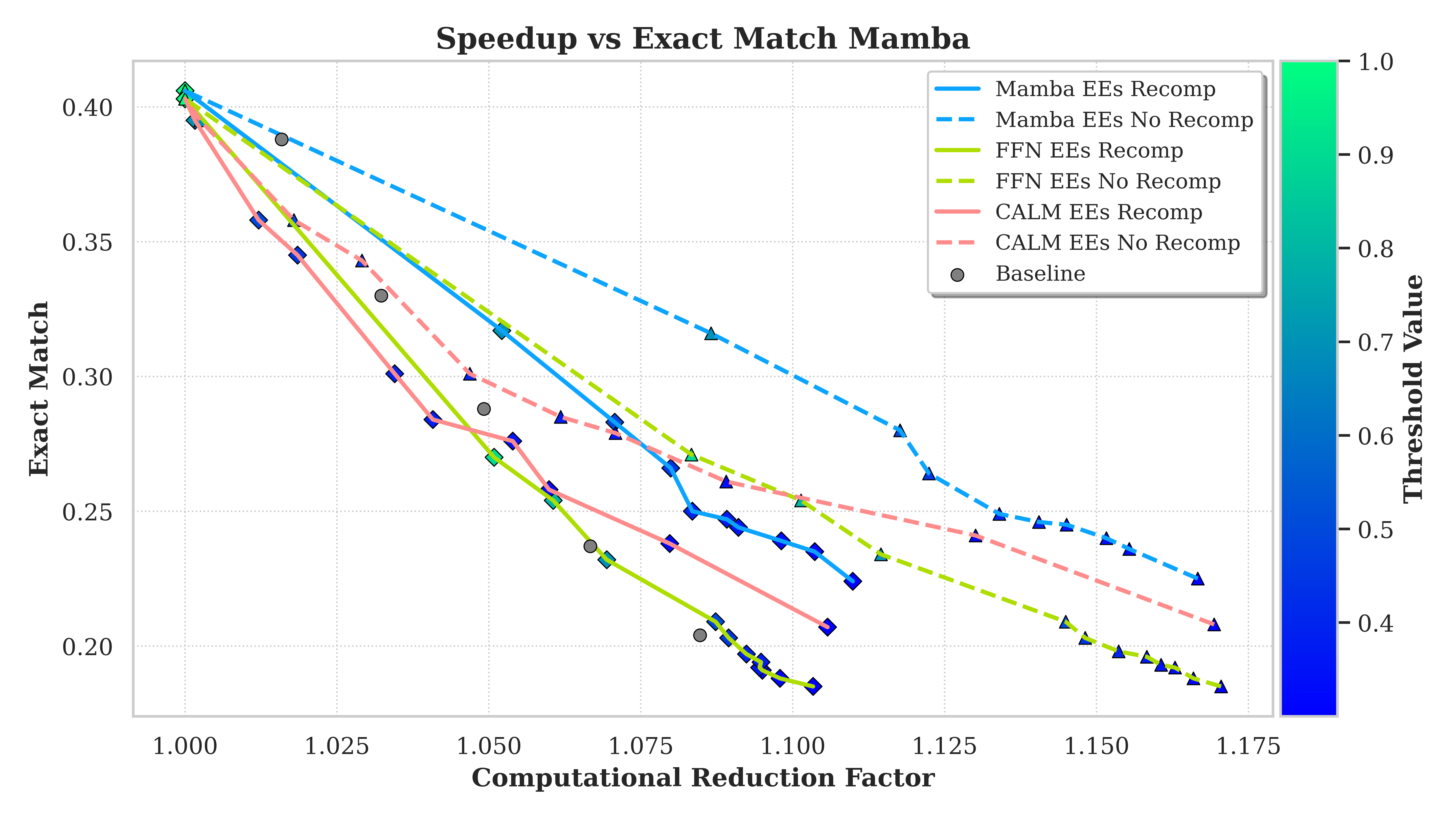}
        \label{fig:mamba_triviaqa}
    \end{subfigure}
    \caption{Comparison of EE and layer pruning performances in the TriviaQA set for Mistral and Mamba 7B models.}
    \label{fig:comparison_triviaqa}
\end{figure}

Figure \ref{fig:comparison_triviaqa} presents the performance results for the TriviaQA task, comparing early exits and layer pruning for the Transformer and Mamba models. Early exits generally outperform layer pruning by achieving a better balance between computational savings and model accuracy. This occurrence is likely to be happening due to having a big information loss when using layer pruning, showing early exits' largest strength, selective usage of computation. The effectiveness of EE is particularly evident at lower confidence thresholds, where gains are modest (around $1.2\times$) but still result in higher accuracy than at baseline. In contrast, layer pruning causes noticeable performance drops, particularly in knowledge-focused tasks such as TriviaQA, where the model does not rely on external information.

The recomputation or not of states appears to have a limited impact on performance improvement, but diverges noticeably between the two architectures. In the Transformer model, the recomputation does not affect too much total computation spent, but it shows that it may allow for more consistent results in some cases, where lower confidence thresholds can be set. The case of Mamba is very different, as it shows a bigger difference between recomputation or not. This event seems to be linked to the higher cost of recomputing the states with the Mamba-based backbone, and it is consistent with the different classifiers.

Between the three configurations of the EE classifier, there is a big difference between the CALM-style classifiers and the other two. These classifiers seem not enough for this problem, compared to their original application. Between the other two, even though of similar parameter sizes, Mamba shows as a better model for EE prediction, possibly because Mamba provides a state, which can keep information of the currently generated text and update it as it goes, which does not happen with the FFN. It also allows for a higher degree of freedom to set the confidence threshold, with a larger spawn over the x axis for the Transformer.

When comparing both model backbones, the EEs provide similar performance gains relative to the baseline, with a consistent degradation per computational cost reduction.

\subsubsection{CoQA}

\begin{figure}[h]
    \centering
    \includegraphics[width=0.5\textwidth, trim={0 0 0 0}, clip]{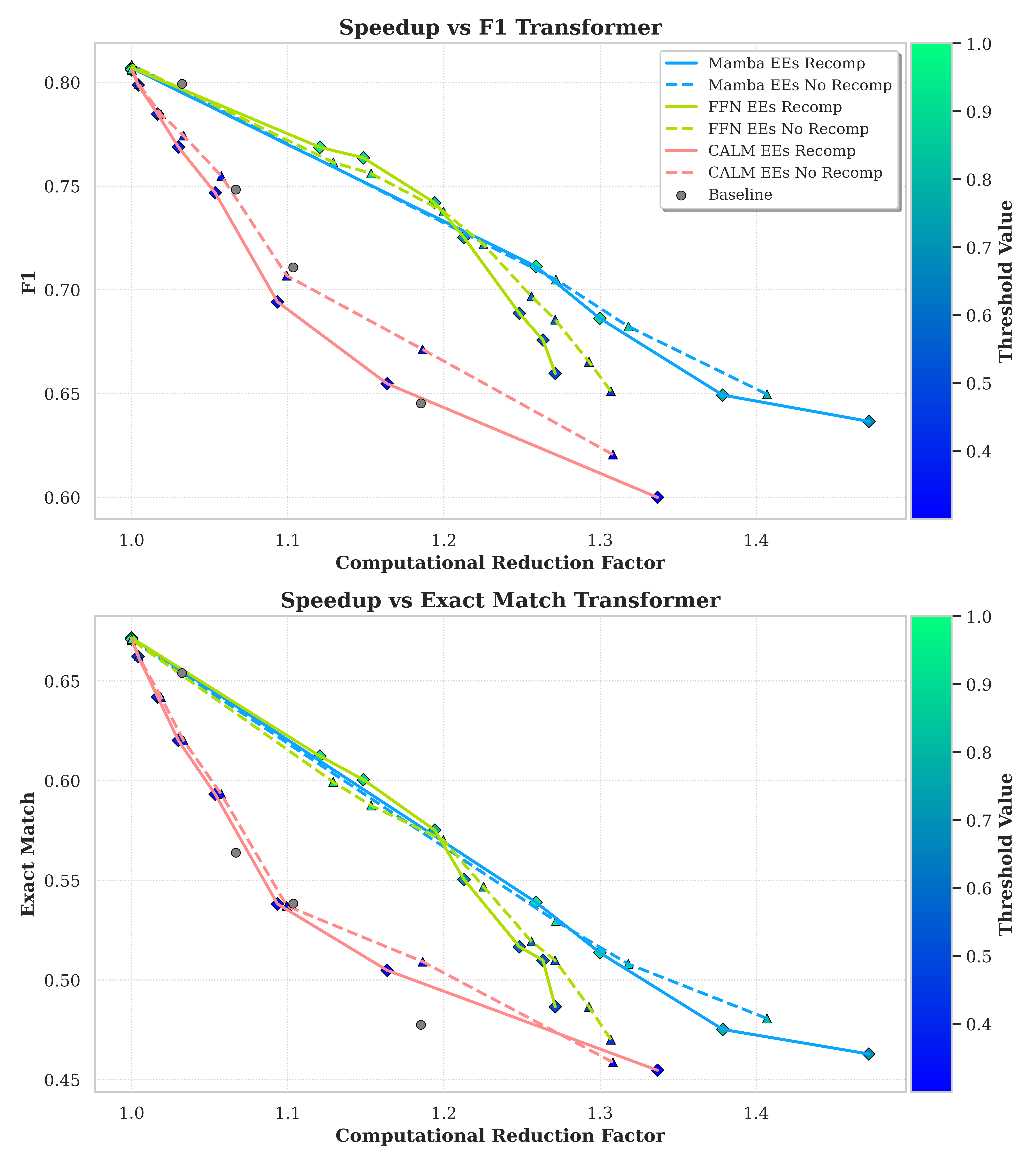}
    \caption{Performance of EE and layer pruning in the CoQA set for Mistral 7B.}
    \label{fig:mistral_coqa}
\end{figure}

\begin{figure}[h]
    \centering
    \includegraphics[width=0.5\textwidth, trim={0 0 0 0}, clip]{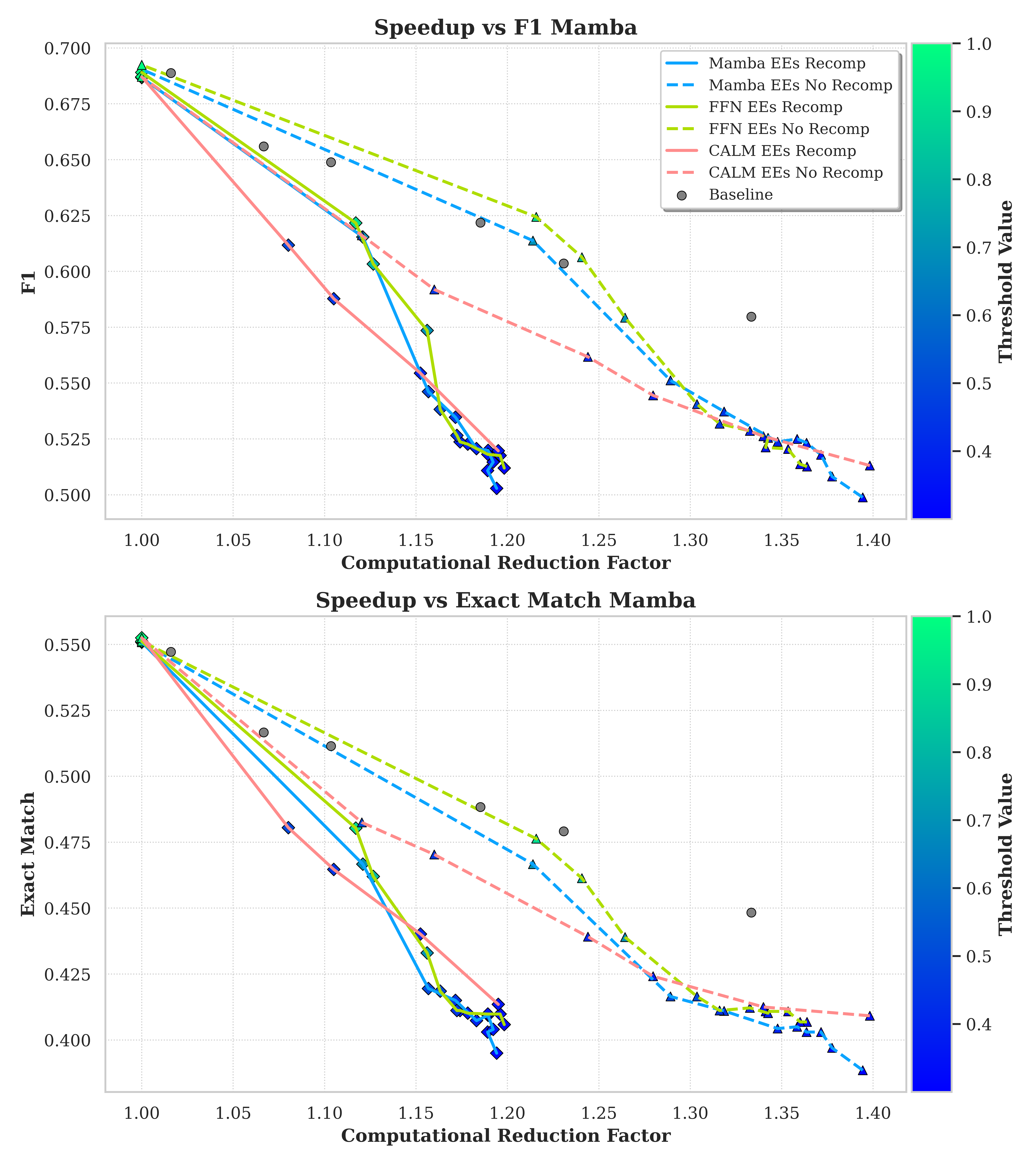}
    \caption{Performance of EE and layer pruning in the CoQA set for Mamba.}
    \label{fig:mamba_coqa}
\end{figure}

The results on CoQA, displayed in the Figure \ref{fig:mistral_coqa} for Transformer and in Figure \ref{fig:mamba_coqa} for the Mamba based architecture, reveal a similar trend to TriviaQA, but with layer pruning being more interesting overall. The Transformer works great in general with the EE, while Mamba does not reach that level of performance with the exits, showing that this pruned configuration works well.

Again, not recomputing the states does not show a significant improvement over recomputing them in the Transformer case, unlike Mamba which shows a big difference. The behaviours of the models both considering Exact match or F1 as a metric are very similar, but the difference in F1 is a bit greater in the case of the recomputation vs no recomputation. Exact match might be a bit strict, so having a better performance with the F1 metric means that even though the model is not outputting exactly the same tokens, they still keep a similar meaning.

The trend of the CALM classifiers is very similar to the one presented by the layer pruning and seems to be the opposite of the FFN. However, the latter is still better computationally. Using a Mamba based classifier seems best mainly thanks to the greater range of confidence thresholds that are available. 

However, for higher computational savings, EE configurations provide better performance. In particular, the CoQA task benefits more from not recomputing states, especially as shown by the F1 metric, which indicates that recomputation may introduce unnecessary overhead without proportionate gains in performance.

\subsubsection{TruthfulQA generation}

\begin{figure}[h]
    \centering
    \includegraphics[width=0.5\textwidth, trim={0.5cm 0 0 0}, clip]{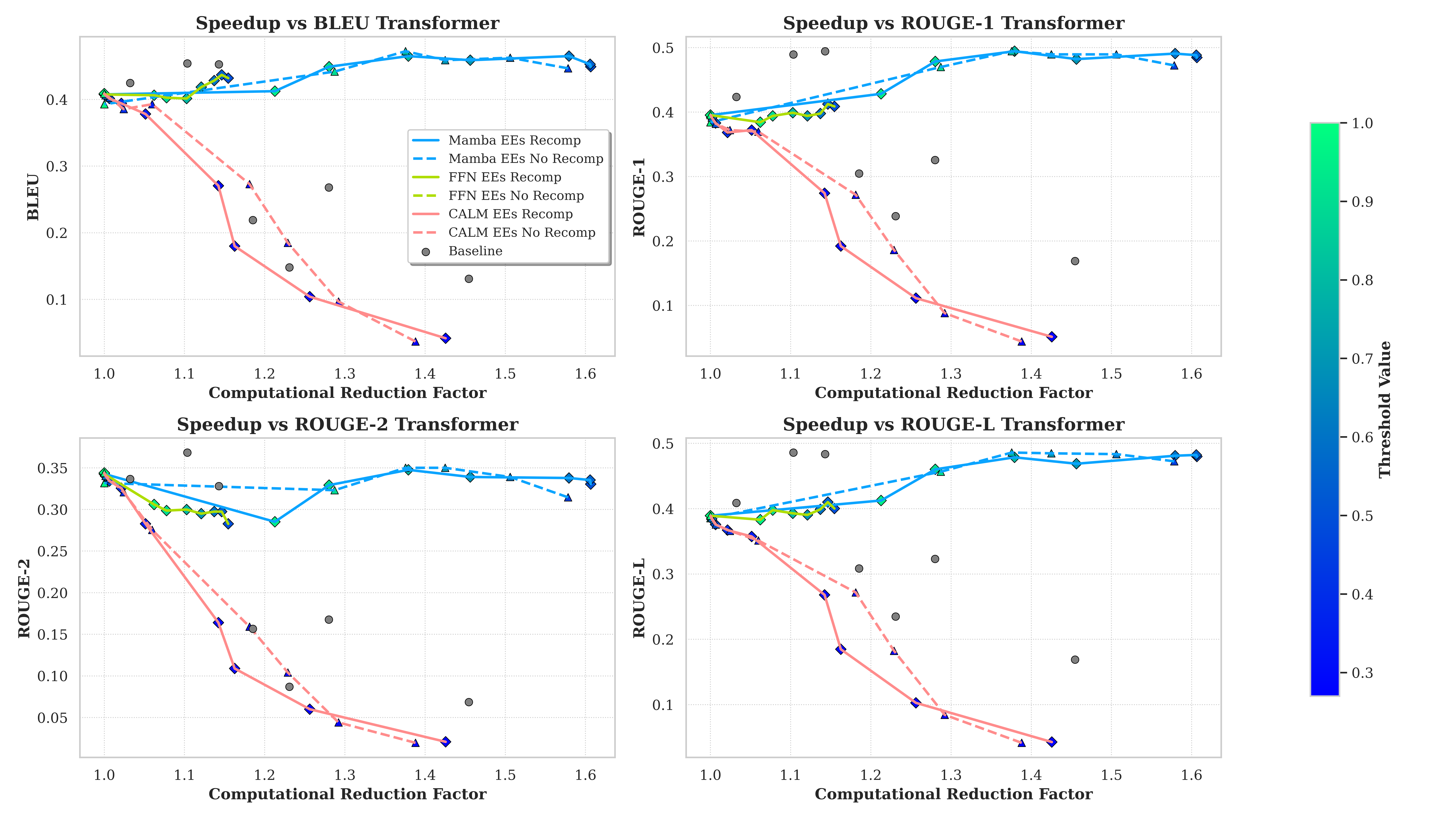}
    \caption{Performance of EE and layer pruning in the Truthful QA generation (acc) set for Mistral 7B.}
    \label{fig:mistral_truthfulqa_max}
\end{figure}

\begin{figure}[h]
    \centering
    \includegraphics[width=0.5\textwidth, trim={0.5cm 0 0 0}, clip]{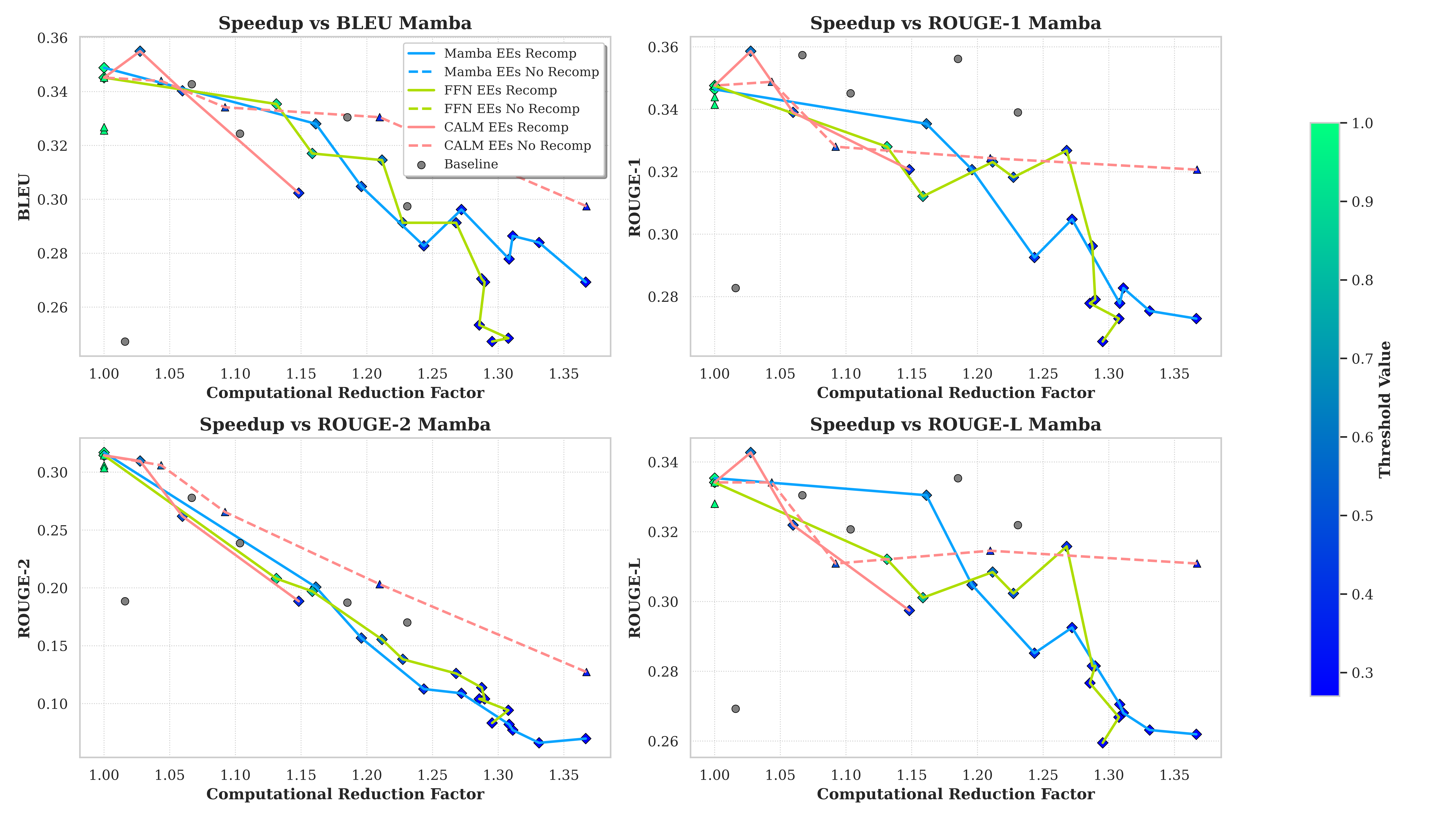}
    \caption{Performance of EE and layer pruning in the Truthful QA generation (acc) set for Mamba.}
    \label{fig:mamba_truthfulqa_max}
\end{figure}

The performance for the TruthfulQA generation task, shown in Figures \ref{fig:mistral_truthfulqa_max} and \ref{fig:mamba_truthfulqa_max} presents a more complex behavior than the one presented in the previous tasks, probably because this task is more focused on the text generation aspect than on the knowledge part, both due to the task itself and the metrics that are used to account for its performance.

For example, in general, the EE configuration is more interesting than the layer pruning strategy, mostly with the Transformer, and slightly less so with the Mamba model. The latter presents a big drop in performance while in Transformer the performance is kept or even improved. Improvements in performance are likely due to a combination of a maintenance in the possible answers of the model, and at the same time a drop in the wrong answers. 

Analyzing the performance of the individual EE configurations, the Mamba based configuration shows a great improvement in performance over the other two, achieving a great computational reduction factor with no performance loss in the task and with a larger range of freedom compared to the other alternatives.

This task seems to be more suited for the analysis of the Transformer model, which excels. This could be partially due to the set of weights selected for the tests.

For this analysis, we used the \textit{acc} submetric of TruthfulQA, which exhibited the most interesting performance patterns, among the three submetrics. Unlike exact match metrics, which are rigid, Bleu and Rouge offer a more flexible evaluation that captures nuanced differences in generated text.

\section{Conclusions and Future Work}

This work investigated early exit mechanisms (EEs) in large language models (LLMs), specifically within the Transformer and Mamba architectures, to improve computational efficiency and inference speed. By implementing EEs in models such as Mistral 7B and Codestral 7B, we assessed the impact of dynamic computation on performance, accuracy, and energy consumption in NLP tasks, using datasets like TruthfulQA, TriviaQA, and CoQA. The results indicate that EEs effectively reduce inference time in low-latency applications, showing a good accuracy-inference time trade-off. In particular, the addition of Mamba-based EE classifiers showed to add more resilience to performance degradation, leveraging their state-space structure to complement early exits, thus supporting memory-efficient processing of long-range dependencies and with a constant inference cost.

Although EEs offer promising enhancements for efficient NLP deployment, they require careful tuning to balance computational savings with model fidelity, particularly in high-accuracy scenarios. Future work could refine EE strategies through adaptive thresholds that respond dynamically to context, latency, or device constraints. Combining EE with optimizations such as pruning and knowledge distillation could further improve performance for resource-constrained applications. Additionally, applying EEs in pretraining, or integrating parameter-efficient fine-tuning methods, may foster confidence-aware models that perform effectively across diverse, high-demand tasks, such as medical diagnostics and autonomous systems, where transparency and efficiency are crucial.







\bibliography{main}
\bibliographystyle{ieeetr}


\end{document}